%% file: acl2021.tex
\documentclass[11pt,a4paper]{article}
\usepackage[hyperref]{acl2021}
\usepackage{times}
\usepackage{latexsym}

\usepackage{microtype}

\usepackage{array}
\usepackage{pifont}
\usepackage{tabularx}
\usepackage{adjustbox}
\usepackage{multirow}
\usepackage{enumitem}
\usepackage{xspace}
\usepackage{tcolorbox}
\usepackage{booktabs,subcaption,amsfonts,dcolumn}
\usepackage{hyperref}
\usepackage{url}
\usepackage{amsmath,amsthm,amsfonts,amssymb,bm,stmaryrd}
\usepackage{xcolor}		
\usepackage{makecell}

\input{header}

\aclfinalcopy % Uncomment this line for the final submission
\def\aclpaperid{988} %  Enter the acl Paper ID here

\title{
Manual Evaluation Matters: Reviewing Test Protocols of\\
Distantly Supervised Relation Extraction

}

\author{Tianyu Gao$^1$, Xu Han$^2$, Keyue Qiu$^2$, Yuzhuo Bai$^2$, Zhiyu Xie$^2$, Yankai Lin$^3$\\\textbf{Zhiyuan Liu$^{2*}$, Peng Li$^3$, Maosong Sun$^2$ and Jie Zhou$^3$}\\
$^1$Department of Computer Science, Princeton University\\
$^2$Department of Computer Science and Technology, Tsinghua University\\
$^3$Pattern Recognition Center, WeChat AI, Tencent Inc\\
{\tt tianyug@princeton.edu}\\
{\tt \{hanxu17, qky18, byz18, xiezy19\}@mails.tsinghua.edu.cn}
}

\date{}

\begin{document}
\maketitle

\renewcommand{\thefootnote}{\fnsymbol{footnote}}
\footnotetext[1]{Corresponding author e-mail: liuzy@tsinghua.edu.cn}
\renewcommand{\thefootnote}{\arabic{footnote}}

\begin{abstract}

\begin{comment}
% Distant supervision can automatically generate large-scale labeled data for relation extraction, 
% and thus many efforts have been devoted into distantly supervised relation extraction (DS-RE). 
% Though with great success, its research has encountered several challenges that have not yet been well addressed or explored: how to credibly evaluate DS-RE models and how to utilize DS-RE data in a more practical way. 
% %the noise problem in evaluation, relationship with pre-training models, and how to better utilize DS-RE data in a more practical way.
% In this paper, we address these questions in three parts:
% (1) We present large human-annotated test sets for two DS-RE datasets to enable more accurate evaluation, and then perform a system-wise comparison in both held-out and human-annotated settings;
% (2) We explore different training strategies for DS-RE, investigating their influence to various systems, including pre-training models;
% (3) We discuss what is the effective way to combine DS-RE data with supervised relation extraction tasks.
% We take extensive experiments to support our views, and we hope these points can further boost the development of DS-RE.
% %We show with our extensive experiments that taking on these new paths is crucial for the development of DS, and there are some new directions worth further exploring in the future.
\end{comment}

Distantly supervised (DS) relation extraction (RE) has attracted much attention in the past few years as it can utilize large-scale auto-labeled data. 
However, its evaluation has long been a problem: previous works either take costly and inconsistent methods to manually examine a small sample of model predictions, 
or directly test models on auto-labeled data---which, by our check, produce as much as 53\% wrong labels at the entity pair level in the popular \nyt~dataset. 
% that can lead to a 53\% error rate of labels at the entity pair level. 
% This problem has not only led to inaccurate evaluation in the history of DS-RE, 
% but also made it hard to understand what the current success is and what the future direction is. 
This problem has not only led to inaccurate evaluation, but also made it hard to understand where we are and what's left to improve in the research of DS-RE.
To evaluate DS-RE models more credibly, 
% we build two manual-evaluated benchmarks \nyt~and \wiki~instead of conventional automatic ones, and thoroughly evaluate several competitive models, especially the latest pre-trained ones. 
we build manually-annotated test sets for two DS-RE datasets, \nyt~and \wiki, and thoroughly evaluate several competitive models, especially the latest pre-trained ones. 
The experimental results show that the manual evaluation can indicate very different conclusions from automatic ones, especially some unexpected observations, e.g., pre-trained models can achieve dominating performance while being more susceptible to false-positives compared with previous methods. 
We hope that both our manual test sets and observations can help advance future DS-RE research.\footnote{Our code and data are publicly available at \url{https://github.com/thunlp/opennre}.}
% and emphasize the opportunities and challenges brought by pre-trained models to DS-RE.
% For example, our annotated test sets show that though achieving supreme performance, pre-trained models are more susceptible to false-positives than previous ones, posing new challenges to DS-RE research.
%the opportunities and challenges brought by pre-trained models to DS-RE.

\end{abstract}

\input{sections/intro.tex}

\input{sections/related.tex}
\input{sections/dataset.tex}
\input{sections/model.tex}

\input{sections/exp.tex}
% \input{sections/discussion.tex}

% \section{Discussion}
% 
% \tianyu{Working in progress of this section.}
% 
% By introducing two large human-labeled test sets, we have not only provided 
% 
% From these results, we can find that \textbf{how to adopt pre-trained models to handle DS data (especially handle large amounts of noisy instances) still remains an open problem.}
% 
% Hence, \textbf{we advocate that later works should also include macro F1 for more comprehensive comparisons.}
% 
% 
% \textbf{These results show that it is promising to distinguish N/A instances into the instances without any relations and the ones with some specific unknown relations. }

\section{Conclusion}

In this paper, we study the problem of test protocols in DS-RE and build large manually-annotated test sets for two DS-RE datasets, to enable a more accurate and efficient evaluation. We not only demonstrate that our manual test sets show different observations from previous held-out ones, but also capture some interesting reflections by using the manual test, e.g., pre-trained models suffer false-positives more and bag-level training strategies generally do not help with pre-trained models. 

We hope that our manual test sets can mark a new starting line for DS-RE, while these observations can motivate novel research directions towards better DS-RE models, e.g., studying denoising methods for pre-trained models or processing N/A relations in a more fine-grained way.

\section*{Acknowledgments}

This work is supported by the National Key Research and Development
Program of China (No. 2020AAA0106501) and Beijing Academy of
Artificial Intelligence (BAAI).
This work is also supported by the Pattern Recognition Center, WeChat AI, Tencent Inc.

\section*{Ethical Considerations}
 
Our work mainly focuses on two parts: the construction of two manually-labeled test sets and the analyses of models based on the manual test. Regarding the annotation, we first approximate the workload by annotating a few examples on our own, and then determine the wages for annotators according to local standards. The two datasets are based on NYT and Wikipedia, and we did not identify any unethical content during annotation. 

Concerning the analyses, we find that models tend to utilize some shallow clues for classification, such as learning heuristics from entities. This behavior can potentially create biased extraction results based on the distributions of entities in the training set and is worth further investigating. 

\bibliographystyle{acl_natbib}
\bibliography{acl2021}

\clearpage
\counterwithin{figure}{section}
\counterwithin{table}{section}
\appendix

\input{sections/appendix.tex}

\end{document}

%% file: header.tex
\providecommand{\todo}[1]{
    {\protect\color{red}{[TODO: #1]}}
}

\providecommand{\tianyu}[1]{
    {\protect\color{blue}{[Tianyu: #1]}}
}

\providecommand{\lyk}[1]{
    {\protect\color{red}{[Yankai: #1]}}
}

\newcommand\tf[1]{\textbf{#1}}
\newcommand\ttt[1]{\texttt{#1}}

\newcommand\nyt{NYT10}
\newcommand\wiki{Wiki20}
\newcommand\idt{\quad}

%% file: sections/intro.tex
%!TEX root = ../acl2021.tex

\section{Introduction}

Relation extraction (RE) aims at extracting relational facts between entities from the text. One crucial challenge for building an effective RE system is how to obtain sufficient annotated data. To tackle this problem, \citet{mintz2009distant} propose distant supervision (DS) to generate large-scale auto-labeled data by aligning relational facts in knowledge graphs (KGs) to text corpora, with the core assumption that one sentence mentioning two entities is likely to express the relational facts between the two entities from KGs. 

\begin{figure}[t]
    \centering
    \includegraphics[width=0.49\textwidth]{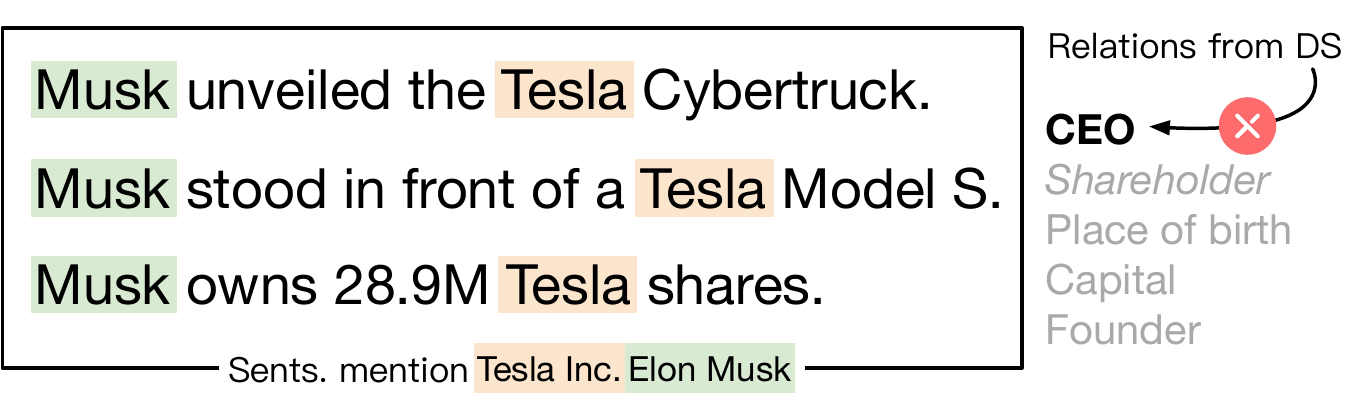}
    %\vspace{-10pt}
    \caption{Typical errors made by DS evaluation. In the figure, DS labels the bag with only the relation \ttt{CEO}, while none of the sentences express the relation. Also, it misses a correct relation \ttt{shareholder} due to the incompleteness of the knowledge graphs.}
    \vspace{-10pt}
    \label{fig:ds}
\end{figure}

As DS can bring hundreds of thousands of auto-labeled training instances for RE without any human labor, DS-RE 
has been widely explored in the past few years~\cite{riedel2010modeling,hoffmann2011knowledge,zeng2015distant,lin2016neural,feng2018reinforcement,vashishth2018reside} and has also been widely extended to other related domains, such as biomedical information extraction~\cite{peng2017cross,quirk2017distant} and question answering~\cite{bordes2015large,chen2017reading}.

%\tianyu{Probably put a figure, explaining different ways of evaluation?}

Although DS-RE has achieved great success, we identify one severe problem for the current DS-RE research---its evaluation. 
Existing works usually take two kinds of evaluation methods following~\citet{mintz2009distant}:
\emph{held-out evaluation}, which directly uses DS-generated test data  to approximate the trend of model performance, and 
\emph{human evaluation}, which manually checks the most confident relational facts predicted by DS-RE models. 
%To get the trend of model performance in an automatic way, held-out evaluation uses the distant supervision test data to measure the quality of extracted relational facts. 
%Human evaluation, in a further step, verifies model predictions by manual checking. 
% Although much more accurate, 
%human evaluation is more accurate than held-out one, it is more costly. 
% human evaluation is too costly compared to the held-out one. 
Since manually checking is costly, most works with human evaluation only examine a small proportion of the predictions. Moreover, different works may sample different splits of data, making human evaluation inconsistent across the literature.
% Most existing works with human evaluation~\cite{mintz2009distant,riedel2010modeling,feng2018reinforcement,jiang2018revisiting,shahbazi2020relation} take only a small portion of their models' predictions for manual evaluation, making it inconsistent and not comparable to other works. 
Most recent studies even skip the human evaluation for the above factors and only take the held-out one.

However, the held-out evaluation can be quite noisy: there are many false-positive cases, where the sentences do not express the auto-labeled relations at all; besides, due to the incompleteness of KGs, 
DS may miss some relations, just as shown in Figure~\ref{fig:ds}. 
% entity pairs with a positive relation might be wrongly labeled as ``not applicable (N/A)''. 
After checking $9,744$ sentences in the held-out set of \nyt~\cite{riedel2010modeling}, the most popular DS-RE dataset, we found that about $53\%$ of the entity pairs are wrongly labeled, emphasizing the need for a more accurate evaluation.

To make DS-RE evaluation more credible and alleviate the trouble of manual checking for later work, 
%we build two manual-evaluated benchmarks based on two popular DS-RE datasets: 
we build human-labeled test sets for two DS-RE datasets: 
\nyt~\cite{riedel2010modeling} and \wiki~\cite{han2020data}.
For \nyt, we manually annotate sentences with positive DS relations in its held-out test set. 
We also use a fine-tuned BERT-based~\cite{devlin2019bert} RE model to predict all ``N/A'' (not applicable) sentences, and manually label the top $5,000$ sentences scored as having a relation. 
Additionally, we merge some unreasonably split relations and reduce the number of relation types from $53$ to $25$. 
For \wiki~dataset, we utilize both the relation ontology and human-labeled instances of an existing supervised dataset Wiki80~\cite{han-etal-2019-opennre} for the test, 
% and align the KG Wikidata\footnote{https://www.wikidata.org/wiki/Wikidata:Main\_Page} and the corpora Wikipedia\footnote{https://en.wikipedia.org/wiki/Main\_Page} to auto-labeled the training set of \wiki.
and then re-organize the DS training data accordingly. 

Based on the newly-constructed benchmarks, 
we carry out a thorough evaluation of existing DS-RE methods, 
as well as incorporating recently advanced pre-trained models like BERT~\cite{devlin2019bert}. 
We found that our manually-annotated test sets can indicate very different conclusions from the held-out one, especially with some surprising observations: 
(1) although pre-trained models lead to large improvements, they also suffer from false-positives more severely, probably due to the pre-encoded knowledge they have~\cite{petroni-etal-2019-language};
(2) existing DS-RE denoising strategies that have been proved to be effective generally do not work for pre-trained models, suggesting more efforts needed for DS-RE in the era of pre-training. To conclude, our main contributions in this work are:
% \begin{itemize}
%     \item We provide two benchmarks with large human-labeled test sets based on widely-used DS-RE datasets, making it possible for later DS-RE works to evaluate in an accurate and efficient way.
%     \item We thoroughly study previous DS-RE methods using both held-out and human-labeled test sets, and find that human-labeled data can reveal inconsistent results compared to the distant ones.
%     \item We discuss some novel and important observations revealed by manual evaluation, especially
%     pre-trained models have dominating performance in DS-RE while exhibiting that they largely suffer from the false positive problem, and existing denoising methods do not work well for pre-trained ones, calling for more research in these directions.
% \end{itemize}
\vspace{-5pt}
\begin{itemize}[leftmargin=*, noitemsep]
    \item We provide large human-labeled test sets for two DS-RE benchmarks, making it possible for later work to evaluate in an accurate and efficient way. %\vspace{3pt}
    \item We thoroughly study previous DS-RE methods using both held-out and human-labeled test sets, and find that human-labeled data can reveal inconsistent results compared to the held-out ones. %\vspace{3pt}
    \item 
    We discuss some novel and important observations revealed by manual evaluation, 
    especially with respect to pre-trained models, which %pre-trained models have dominating performance in DS-RE while exhibiting that they largely suffer from the false positive problem, and existing denoising methods do not work well for pre-trained ones, 
    calls for more research in these directions. 
\end{itemize}
\vspace{-3pt}

%% file: sections/related.tex
%!TEX root = ../acl2021.tex

\input{tables/dataset.tex}

\section{Related Work}
\label{sec:related}

Relation extraction is an important NLP task and has gone through significant development in the past decades. 
In the early days, RE models mainly take statistical approaches, such as 
pattern-based methods~\cite{huffman1995learning,califf-mooney-1997-relational}, 
% feature-based methods~\cite{kambhatla2004combining,guodong2005exploring,jiang2007systematic,nguyen2007relation}, 
feature-based methods~\cite{kambhatla2004combining,guodong2005exploring}, 
%graphical methods~\cite{roth2002probabilistic,roth2004linear,sarawagi2005semi,yu2010jointly}, etc.
graphical methods~\cite{roth2002probabilistic}, etc.
%With the increasing computing power and the development of deep learning, neural RE methods have shown a great success~\cite{liu2013convolution,zeng2014relation,zhang2015relation,nguyen2015combining,vu2016combining,zhang2018graph,zhu2019graph}. 
With the increasing computing power and the development of deep learning, neural RE methods have shown a great success~\cite{liu2013convolution,zeng2014relation,zhang2015relation,zhang2017position}. Recently, pre-trained models like BERT~\cite{devlin2019bert} have dominated various NLP benchmarks, including those in RE~\cite{soares2019matching,zhang2019ernie}. All these RE methods focus on training models in a supervised setting and require large-scale sufficient human-annotated data.

To generate large-scale auto-labeled data without human effort, \citet{mintz2009distant} first use DS to label sentences mentioning two entities with their relations in KGs, which inevitably brings wrongly labeled instances. 
To handle the noise problem, \citet{riedel2010modeling,hoffmann2011knowledge,surdeanu2012multi} apply multi-instance multi-label training in DS-RE. % and \citet{hoffmann2011knowledge,surdeanu2012multi} take a multi-instance multi-label training. 
Following their settings, later research mainly takes on two paths: 
one aims at selecting informative sentences from the noisy dataset, using heuristics~\cite{zeng2015distant}, 
attention mechanisms~\cite{lin2016neural,han2018hierarchical,zhu2019graph}, 
adversarial training~\cite{wu2017adversarial,wang2018adversarial,han2018denoising}, 
and reinforcement learning~\cite{feng2018reinforcement,qin2018robust};
the other incorporates external information like 
KGs~\cite{ji2017distant,han2018neural,zhang2019long,hu2019improving}, multi-lingual corpora~\cite{verga2016multilingual,lin2017neural,wang2018adversarial}, 
as well as relation ontology and aliases~\cite{vashishth2018reside}. 
Recently, pre-trained DS-RE models have also been explored, including both domain-general~\cite{alt2019fine,xiao2020joint} and domain-specific~\cite{amin2020data} models.
Some other latest works~\cite{peng2020learning} utilize DS data for intermediate pre-training in order to boost supervised RE tasks.

As mentioned in our introduction, the evaluation of DS-RE has long been a problem, especially since many existing methods solely rely on auto-labeled test data. Some preliminaries have noticed this problem: 
\citet{jiang2018revisiting,zhu-etal-2020-towards-accurate} also annotate the test set of \nyt, yet \citet{jiang2018revisiting} only sample $2,040$ sentences from it, and \citet{zhu-etal-2020-towards-accurate} discard all N/A data from DS, which are an important part of DS evaluation, and assume that the original held-out data have either the DS relations or no relation at all, while we find that a large proportion of held-out data actually express some other relations; 
\citet{li2020active-testing} propose active testing, an iterative method to correct the bias of DS evaluation. However, it still requires consistent human efforts during each evaluation phase. 
To the best of our knowledge, our work, 
building benchmarks with large-scale manually-labeled test data, conducts the most comprehensive human evaluations of DS-RE methods so far.

%% file: tables/dataset.tex
%!TEX root = ../acl2021.tex

\begin{table*}[t]
    \small
    \begin{center}
    \centering
    %\resizebox{2.0\columnwidth}{!}{%
    \begin{tabular}{l|r|rrr|rrr|rrr}
    \toprule
    \multirow{2}{*}{Dataset} & \multirow{2}{*}{\#rel} & \multicolumn{3}{c|}{Train} & \multicolumn{3}{c|}{Validation} & \multicolumn{3}{c}{Test}\\
     & & \#facts & \#sents & N/A & \#facts & \#sents & N/A & \#facts & \#sents & N/A\\
     \midrule
    %  \nyt$^{\dagger}$ & 53 & 18,409 & 522,611 & 74\% & - & - & - & 1,950 & 172,448 & 96\%\\
    %  \nyt & 25 & 17,137 & 417,893 & 80\% & 4,062 & 46,422 & 80\% & \makecell[r]{3,899\vspace{-2pt}\\{\scriptsize(1,940)}} & \makecell[r]{9,744\vspace{-2pt}\\{\scriptsize(157,859)}}  & \makecell[r]{32\%\vspace{-2pt}\\{\scriptsize(96\%)}} \\
     \nyt$^{\dagger}$ & 53 & 18,409 & 522,611 & 74\% & - & - & - & 1,950 & 172,448 & 96\%\\
     \nyt & 25 & 17,137 & 417,893 & 80\% & 4,062 & 46,422 & 80\% & 3,899 & 9,744 & 32\%\vspace{-2pt}\\
&&&&&&&&{\scriptsize(1,940)}&{\scriptsize(157,859)}&{\scriptsize(96\%)}\\
     \wiki$^{\dagger}$ & 454 & 203,176 & 1,050,246 & 48\% & 4,333 & 29,145 & 48\% & 4,333 & 28,897 & 48\% \\
     \wiki & 81 & 157,740 & 698,721 & 59\% & 17,485 & 64,607 & 73\% & 56,000 & 137,986 & 25\% \\
    \bottomrule
    \end{tabular}
   % }
    \end{center}
    \caption{The statistics of the datasets used for our benchmarks, including both the original ($\dagger$) and our modified versions. 
    We list the numbers of relations (\#rel), relational facts (\#facts) and sentences (\#sents), and N/A rate (N/A) for these datasets. 
    For \nyt, numbers in brackets are for the held-out test, otherwise for bag-level manual test. 
    }
    \label{tab:dataset}
    \vspace{-0.5em}
\end{table*}

%% file: sections/dataset.tex
%!TEX root = ../acl2021.tex

\section{DS-RE Datasets}
\label{sec:dataset}

In this section, we introduce the way we build the manually-annotated test sets for \nyt~\cite{riedel2013relation} and \wiki~\cite{han2020data}.
% and how we construct the human-labeled test sets for them.
% modify them into benchmarks with human-labeled test sets. 
We show the statistics of these datasets in Table~\ref{tab:dataset}.

\subsection{\nyt~Dataset}
\label{sec:dataset-nyt}

\nyt~is constructed by aligning facts from the FreeBase~\cite{bollacker2008freebase} with the New York Times (NYT) corpus~\cite{sandhaus2008nyt}. 
The original \nyt~dataset contains $53$ relations (including N/A). After thoroughly examining the dataset, we found that (1) there are many duplicate instances in the dataset, (2) there is no public validation set, and some previous works directly take the test set to tune the model, and (3) the relation ontology is not reasonable for an RE task. 
Therefore, we first clean the dataset by removing duplicate sentences, split a validation set, and then merge some of the relations as described below.

\paragraph{A New Relation Ontology} 

One example of \nyt's improper relation ontology is the relations related to state/province capitals. There are $12$ such relations in the original dataset, representing region capitals of different countries, while some of these relations even do not have instances in the test set. 
We merge these $12$ relations as one unified relation \ttt{/location/region/capital}, and we also merge $3$ relations related to organization locations as \ttt{/business/location}. 
Besides, we delete relations that only show up in either the training set or the test set (most of which have very few sentences).
In the end, we get $25$ relations in the new dataset. We provide the detailed relation list of the new dataset in Appendix~\ref{sec:nyt_rel}.

Interestingly, we found that training models with the new dataset leads to a slight performance degeneration, as shown in Table~\ref{tab:old_new_nyt}, which is counter-intuitive (since merging classes usually makes the task easier).
We suspect that the original relation ontology provides heuristics for the model. 
For example, models can learn from the original ontology that every sentence with a ``US state'' as the head entity expresses the fine-grained relation \ttt{/location/us\_state/capital}, which is a shortcut that cannot be acquired with the merged relation \ttt{/location/region/capital}. 
This shows the bias of the original \nyt~dataset being inappropriately exploited by models.

\paragraph{Annotated Test Set}

The original \nyt~dataset only provides an auto-labeled test set for the held-out evaluation. 
Based on the original test set, we manually annotate all sentences that have a positive DS label. %are not labeled as N/A. 
In addition to that, we also fine-tune a BERT model (as described in \S\ref{sec:model_baselines}) to predict the relations of all sentences originally labeled as N/A, 
and annotate the $5,000$ sentences with the highest predicted scores of non-N/A relations. 
%In the end we take 9,744 sentences for annotation.
% In the annotation, 
For each sentence, we ask $4$ annotators to decide whether it expresses one or more relations among $25$ relations. 
Note that one sentence may have multiple relations.
Specially, we do not provide any relation suggestions using DS labels or model predictions to annotators, in order to avoid the annotators being biased by the external information.

When aggregating the final annotation results, we consider each relation for each sentence independently. 
The sentence is regarded to express one relation if more than half of the annotators labeled it with this relation. 
If one sentence gets no votes for any positive relations in the above process, then it is labeled as N/A. 
For annotation conflicts (i.e., no candidate relation gets more than half votes), the authors manually annotate these sentences.

Finally, we obtain the human-labeled test set with $9,744$ sentences, $32\%$ of which are N/A instances. It contains $5,174$ entity pairs and $3,899$ manually-verified relational facts in total. 
After comparing it with the corresponding original DS-generated labels, 
we found that at the fact level, the DS annotations only have a precision of 69.1\% and a recall of 33.9\%. At the entity pair level, the accuracy of DS labels is only 47.1\%. This emphasizes the need to take the human-annotated test set for more accurate evaluation in DS-RE.

\input{tables/old_new_nyt.tex}

\subsection{\wiki~Dataset}

\citet{han2020data} construct the \wiki~dataset by aligning the English Wikipedia corpus with Wikidata~\cite{vrandevcic2014wikidata}. 
To provide an annotated test set for it, we utilize Wiki80~\cite{han-etal-2019-opennre}, a supervised RE dataset with $80$ relations from Wikidata. 
We re-organize \wiki~by adopting the same relation ontology as Wiki80 and re-splitting the train/validation/test sets, while taking sentences in Wiki80 as the test set. 
 %the test set is composed of both non-N/A and N/A sentences. 
We make sure that there is no overlap of entity pairs among the three splits, to avoid any information leakage.

% Note that \wiki~is very different from \nyt, for its N/A instances can actually be those sentences annotated by DS as other relations outside the relation ontology in \wiki, while in \nyt, entity pairs of N/A sentences do not have any relation in the KG. In other words, N/A sentences in \wiki~tend to indicate one undefined relation, yet N/A sentences of \nyt~are more likely to express no relations between the entities.

Note that \wiki~is quite different from \nyt: \nyt~labels a sentence as ``N/A'' if the entity pair in the sentence does not have a relation in the KG. On the contrary, \wiki~labels entity pairs with a relation outside its relation ontology as ``N/A''. In other words, ``N/A'' instances in \wiki~express \emph{unknown} relations instead of \emph{no} relation.

%% file: tables/old_new_nyt.tex
%!TEX root = ../acl2021.tex

\begin{table}[t]
    \small
    \begin{center}
    \centering
    %\resizebox{1.0\columnwidth}{!}{%
    \begin{tabular}{l|cc|cc}
    \toprule
    \multirow{2}{*}{Model} & \multicolumn{2}{c|}{Original} & \multicolumn{2}{c}{New}\\
    & AUC & F1 & AUC & F1 \\
    \midrule
    % 0.5 micro PCNN+AVG & 31.8 & 37.7 & 28.4 & 35.8 \\
    % 0.5 micro PCNN+ATT & 33.6 & 40.1  & 32.2 & 39.1\\
    PCNN+bag+AVG & 31.8 & 38.6  & 28.4 & 35.8 \\
    PCNN+bag+ATT & 33.6 & 40.2  & 32.2 & 39.1\\
    \bottomrule
    \end{tabular}
    %}
    \end{center}
    \caption{AUC and micro F1 (\%) of model predictions on the original \nyt~relation ontology and our new \nyt~relation ontology (using held-out test). }
    \label{tab:old_new_nyt}
    \vspace{-0.7em}
\end{table}

%% file: sections/model.tex
%!TEX root = ../acl2021.tex

\section{DS-RE Models}
\label{sec:model}

%\todo{}
In this section, 
we elaborate the multi-instance multi-label framework for DS-RE, and introduce models we evaluate, including both previous methods and the latest pre-trained ones.
%and describe how we use pre-trained models in DS-RE.

\subsection{Multi-instance Multi-label Evaluation}

\begin{figure}[t]
    \centering
    \includegraphics[width=0.48\textwidth]{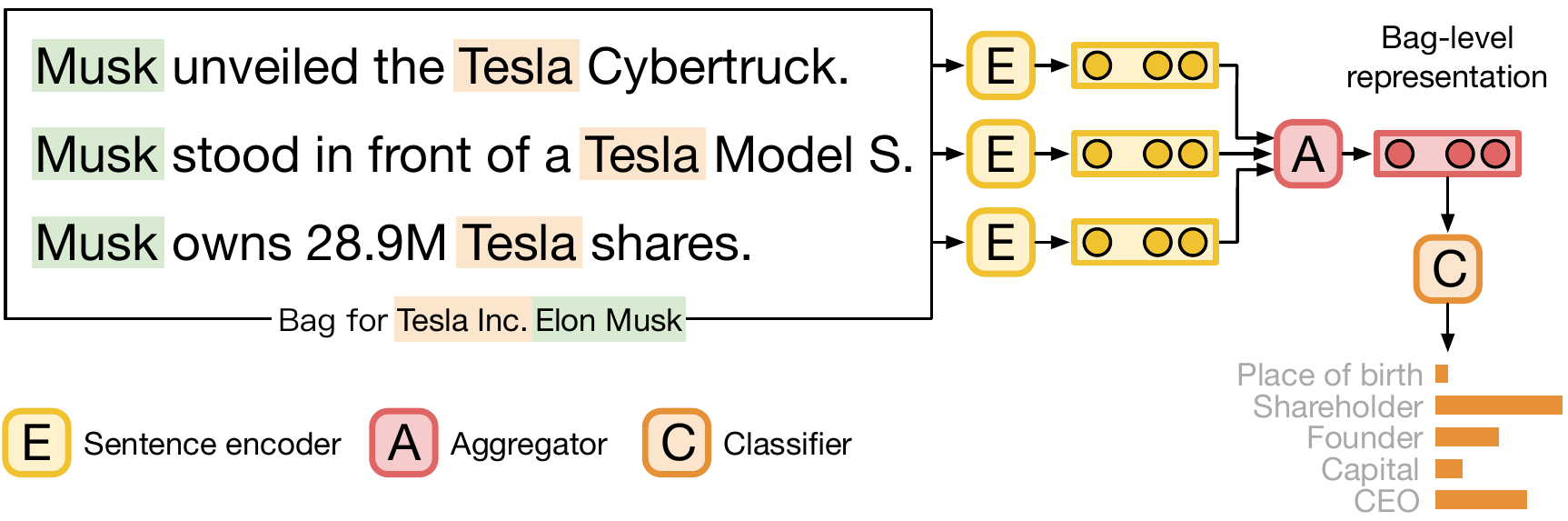}
    %\vspace{-10pt}
    \caption{An illustration for a typical multi-instance multi-label (bag-level) model. The model aims to predict relation probabilities for entity pairs, instead of sentences, which is usually accomplished by aggregating a bag representation and doing classification over it. 
    %This can be adopted solely in testing (which means the model is sentence-level trained) or in both training and evaluation (bag-level training).
    }
    \label{fig:baglevel}
    \vspace{-0.5em}
\end{figure}

Unlike the supervised RE tasks which usually evaluate models at the sentence level, DS-RE evaluates \emph{how well models can extract relational facts} from the corpus, 
i.e. measuring the precision and recall of extracted relational facts (a \emph{fact} is an entity pair and a relation between them). 
It is named as \tf{multi-instance multi-label}, since each entity pair might be mentioned in multiple sentences, and one entity pair can have more than one relation. 
 %rather than from one single sentence. 
%Thus, the evaluation is at the relational fact level (i.e., measuring the precision and recall of extracted relational facts). 
Under the framework, models are required to predict the potential relations for each entity pair---according to all sentences mentioning the pair---during the evaluation, as shown in Figure~\ref{fig:baglevel}. 
%Since each entity pair might be mentioned in multiple sentences, and one entity pair can have more than one relation, this framework is named as multi-instance multi-label. 
%Intuitively, we require models to predict relations of the given entity pair according to all sentences mentioning these entities. 
Sentences correlated with the same entity pair are also named as a \emph{bag}, and thus we interchangeably refer to multi-instance multi-label framework as \tf{bag-level} framework.

The most popular way to compare DS-RE models is to plot precision-recall (P-R) curve and calculate the area-under-the-curve (AUC). We also report micro F1 and macro F1 in our experiments. Since the numbers of instances for different relations are extremely imbalanced, macro F1 can better demonstrate model performance while avoiding the bias brought by those major relations.

\subsection{Model Details}
\label{sec:model_baselines}
% We now introduce two ways of training the model---sentence-level and bag-level training.
%In our experiments, we compare two types of models---the sentence-level and bag-level training.

\paragraph{Sentence-level Training} 
Models are trained in a sentence-level fashion (as in supervised RE), but used as a bag-level model during evaluation. As shown in Figure~\ref{fig:baglevel}, a typical bag-level model takes an \emph{aggregator} to fuse embeddings of all sentences in the bag, and then feeds the bag-level representation to the classifier. 
Here we take two aggregation strategies: 
%These DS-RE models are simply trained in a sentence-level fashion as in supervised RE. To evaluate them under the multi-instance multi-label framework, we take two aggregation methods: 
average (\tf{AVG}), 
which averages the representations of all the sentences in the bag; 
%and feeds the averaged result for classification; 
and at-least-one (\tf{ONE})~\cite{zeng2015distant}, which first predicts relation scores for each sentence in the bag, and then takes the highest score for each relation.

%Average first gets a representation for the entity pair by taking the mean of representations of sentences, which mention the entity pair, 
%and then it takes the entity-pair-representation for classification.
%At-least-one, on the other hand, predicts relation probabilities on a sentence basis, and then for each relation, it takes the highest score among all sentences of each entity pair, as decribed in the following,

% \begin{equation}
%     P(r|h,t)=\max_{x\in \mathcal{S}(h,t)}{P(r|h,t,x)}, r\in \mathcal{R}
% \end{equation}
% where $h$ and $t$ are head and tail entities respectively, $\mathcal{S}(h,t)$ is the set of sentences that mention $(h, t)$, and $\mathcal{R}$ is the relation set.

\input{tables/sample_bag.tex}

\input{tables/main_table.tex}

\paragraph{Bag-level Training} Directly deploying sentence-level training for DS-RE suffers from the wrong labeling problem: 
one sentence mentioning two entities may not express its auto-labeled relation. 
To alleviate this problem, models can also take bag-level framework in the training, based on the \emph{expressed-at-least-one} assumption~\cite{riedel2010modeling}: 
at least one sentence in the bag expresses the auto-labeled relation. 
Besides the AVG and ONE\footnote{During training, since we have DS label $r$ for the bag, we directly take the sentence embedding that has the highest score for $r$ as the bag-level representation.} strategies mentioned above that can be directly deployed for bag-level training, \citet{lin2016neural} also propose to use sentence-level attention (\tf{ATT}) for aggregation: It produces bag-level representation as a weighted average over embeddings of sentences, and determines weights by  attention scores between sentences and relations. 

\paragraph{} 
For our experiments, we take CNN~\cite{liu2013convolution}, PCNN~\cite{zeng2015distant} and BERT~\cite{devlin2019bert} as options of sentence encoders, 
which are all common choices for neural RE models. 
%All these neural architectures are usually adopted as sentence encoders for RE models.
% and follow hyper-parameters from \citet{han-etal-2019-opennre}. 
We evaluate combinations of different sentence encoders, training policies, and aggregation strategies, e.g., bag-level trained PCNN with ATT aggregator (PCNN+bag+ATT) or sentence-level trained BERT with ONE aggregator (BERT+sent+ONE). 
%For example, in Table~\ref{tab:main}, PCNN (ONE) indicates training PCNN in a sentence-level manner and evaluate it by the ONE policy; PCNN+ONE means training and testing PCNN at bag-level with the ONE policy. 
Besides, we evaluate several representative DS-RE models from literature, namely RL-DSRE~\cite{qin2018robust}, which takes deep reinforcement learning for denoising training instances, BGWA~\cite{jat2017improving}, which takes both word-level and sentence-level attention, and RESIDE~\cite{vashishth2018reside}, which introduces side information like relation aliases to put soft constraints on prediction.

% \begin{itemize}
%     \item PCNN+ATT+RL~\cite{qin2018robust} takes deep reinforcement learning to filter false positive sentences in the dataset, and it can be adopted on top of existing bag-level training methods.
%     \item BGWA~\cite{jat2017improving} is a Bi-GRU based model with both word-level and sentence-level attention mechanism.
%     \item RESIDE~\cite{vashishth2018reside} utilizes side information as entity types and relation aliases to put soft constraints in relation prediction.
% \end{itemize}

%Fine-tuning pre-trained models has shown supreme results on supervised RE~\cite{soares2019matching}, 
%thus we also adopt BERT~\cite{devlin2019bert}, a popular pre-trained model, as one of the sentence encoders. 
%We use OpenNRE~\cite{han-etal-2019-opennre} for most experiments, including both sentence-level and bag-level training. 

For BERT-based sentence encoder, there are some practical challenges when adopting bag-level training: 
in the worst cases, one bag can contain thousands of sentences, which are beyond the capacity of most computing devices due to the large size of pre-trained models. 
To address this issue, we take a random sampling strategy during training: for each bag, we randomly sample $b$ sentences, instead of taking all of them. 
For evaluation, we use the same routine as other non-pre-trained encoders, taking all of the sentences into account (because back propagation is not needed here so the bag can be split into several batches). 
Since this is different from the original bag-level training, we carry out a pilot experiment to examine the effect of the sampled training. 
From Table~\ref{tab:sample_bag}, we can see that our sampling strategy does not significantly hurt the performance of the bag-level training. 
% We compare the performance of PCNN+bag+ATT of both standard bag-level training and sampled training in Table~\ref{tab:sample_bag}, and verify that our sampling strategy does not significantly hurt the performance. 

We also add another variant, BERT-M, in our evaluation. 
We observe from the top predictions of BERT models (Figure~\ref{fig:nyt_corr}) that BERT tends to make false-positive errors for entity pairs that express a relation in the KG but do not have any sentence truly expressing the relation in the data, probably due to that model learns shallow cues solely from entities. Thus, following~\citet{peng2020learning}, we mask entity mentions during training and inference to avoid learning biased heuristics from entities.

%% file: tables/sample_bag.tex
%!TEX root = ../acl2021.tex

\begin{table}[t]
    \small
    \begin{center}
    \centering
    %\resizebox{1.0\columnwidth}{!}{%
    \begin{tabular}{l|ccc}
    \toprule
    Model & AUC & Micro & Macro\\
    \midrule
    PCNN+bag+AVG (full) & 28.4 & 	35.8&	13.3 \\
    PCNN+bag+ATT (full) & 32.2 & 	39.1&	9.5 \\
    PCNN+bag+ATT (sample) &31.8 & 	39.6&	11.8 \\
    \bottomrule
    \end{tabular}
    %}
    \end{center}
    \caption{Comparison between full and sampled bag training  with \nyt~held-out test. 
    For PCNN+bag+ ATT, the sampled bag performs similarly to the full bag, and it still outperforms PCNN+bag+AVG (full). ``Micro'' and ``Macro'' represent micro and macro F1 (\%). %demonstrating that sampled bag is still effective bag-level training.
    }
    \label{tab:sample_bag}
    \vspace{-0.5em}
\end{table}

%% file: tables/main_table.tex
\begin{table*}[t]
    \begin{center}
    \centering
    \small
    %\resizebox{2.0\columnwidth}{!}{%
  \setlength{\tabcolsep}{6pt}
    \begin{tabular}{ccc|ccc|ccc|ccc}
    \toprule
    \multirow{2}{*}{Model}&\multirow{2}{*}{Bag}&\multirow{2}{*}{Strategy} & \multicolumn{3}{c|}{Held-out} & \multicolumn{3}{c|}{Bag-level Manual} & \multicolumn{3}{c}{Sentence-level Manual}\\
     && & AUC      & Micro & Macro & AUC  & Micro & Macro & AUC  & Micro & Macro \\
     \midrule
   \multirow{2}{*}{CNN}&- & AVG & 20.0 & 	30.3&	6.5&			48.7&	50.4&	21.3&			52.0&	53.3&	22.1 \\
    &  -& ONE & 21.2 & 	31.8&	7.2&			50.5&	51.6&	19.8&			52.0&	53.3&	22.1\\
    \midrule
    \multirow{6}{*}{PCNN}&-&AVG & 20.4 & 	30.9&	9.0 &			49.4&	51.6&	22.6&			52.2&	54.3&	23.2\\
    &-&ONE &21.4 & 	31.9&	7.8&			51.1&	52.6&	23.8&			52.2&	54.3&	23.2\\
    %\midrule
    \cmidrule{2-12}
    &\checkmark & {AVG}    &28.4 & 	35.8&	13.3&			52.9&	53.6&	23.5&			56.0 &	55.9&	22.9\\
    &\checkmark & {ONE} & 28.4 & 	36.0&	8.0&			53.4&	54.8&	24.5&			55.5&	56.7&	22.2\\
    &\checkmark & {ATT} &32.2 & 	39.1&	9.5&			56.8&	56.5&	25.5&			57.1&	56.1&	23.6\\
    \midrule
    RL-DSRE &\checkmark & - & 32.6 & 39.5 & 13.4 & 55.1 & 55.9 & 26.4 & 55.6 & 56.1 & 23.9 \\
    BGWA & \checkmark & - & 31.0 & 	37.4 & 	11.6 & 	47.8 & 	54.0 & 	14.1 & 42.2 & 48.9 & 7.2\\
    RESIDE & \checkmark & - & 33.4 & 	40.5& 	16.9 & 	35.8 & 	43.3 & 	10.2 & 43.2 & 47.9 & 19.8 \\
    \midrule
    \multirow{5}{*}{BERT}& -&AVG    &50.5 & 	51.2&	21.6&			60.3&	62.4&	35.3&			63.2&	64.3&	34.1\\
    & -&ONE   &  50.5 & 	51.6&	21.2&			61.3&	62.9&	36.1&			63.2&	64.3&	34.1\\
   % \midrule
   \cmidrule{2-12}
    &\checkmark & {AVG}    &43.0 & 	47.4&	22.4&			56.7&	60.4&	35.7&			60.4&	63.9&	34.6\\
    &\checkmark & {ONE}    &38.5 & 	46.1&	10.9&			58.1&	61.9&	33.9&			61.5&	65.1&	32.1\\
    &\checkmark & {ATT}    & 27.8 & 	37.4&	13.4&			51.2&	54.1&	25.8&			54.2&	57.2&	26.4\\
    \bottomrule
    \end{tabular}
    %}
    \end{center}
    \vspace{-3pt}
    \caption{Results (\%) on \nyt, including the held-out evaluation, bag-level manual evaluation, and sentence-level manual evaluation. The ``bag'' column indicates whether the model uses bag-level training, and the ``strategy'' column shows the bag aggregation policy. We report the AUC, micro F1 (Micro) and macro F1 (Macro) scores.}
    \label{tab:main} 
    \vspace{-0.5em}
\end{table*}

%% file: sections/exp.tex
\section{Experiment}
\label{sec:exp}

\subsection{Implementation Details}

We use the OpenNRE toolkit~\cite{han-etal-2019-opennre} for most of our experiments, including both sentence-level and bag-level training. For CNN and PCNN, we follow the hyper-parameters of \citet{han-etal-2019-opennre}. For BERT, we use pre-trained checkpoint \texttt{bert-base-uncased} for initialization, take a batch size of $64$, a bag size of $4$ and a learning rate of $2\times 10^{-5}$,\footnote{This is determined by a grid search over batch sizes in \{16, 32, 64\} and learning rates in \{1e-5, 2e-5, 5e-5\}.} and train the model for $3$ epochs. 
%following the common practice for fine-tuning BERT~\cite{devlin2019bert}. 
For RL-DSRE, RESIDE and BGWA, we directly use their original implementation. %and just replace the data with ours.

\subsection{Evaluation Settings}

We take three different settings in our experiments: %to cover evaluations using both the DS data and the manually-annotated data:

\tf{Held-out evaluation}: We take the test data of the original DS datasets for evaluation. The trend of this evaluation should be consistent with the reported results in most DS-RE literature.

\tf{Bag-level manual evaluation}: We take our human-labeled test data for bag-level evaluation. Since annotated data are at the sentence-level, we construct bag-level annotations in the following way: For each bag, if one sentence in the bag has a human-labeled relation, this bag is labeled with this relation; if no sentence in the bag is annotated with any relation, this bag is labeled as N/A.

\tf{Sentence-level manual evaluation}: As we wonder how well bag-level-trained models can handle sentence-level predictions, our human-labeled test set is also used for a sentence-level evaluation.

We report AUC, micro F1 and macro F1 for all above evaluation settings. We take the best micro F1 on the P-R curves and use the corresponding threshold for calculating macro F1. Considering the difference between \nyt~and \wiki, we evaluate models on the two datasets respectively.

\subsection{The Results on \nyt}

\begin{figure}[t]
    \centering
    \includegraphics[width=0.4\textwidth]{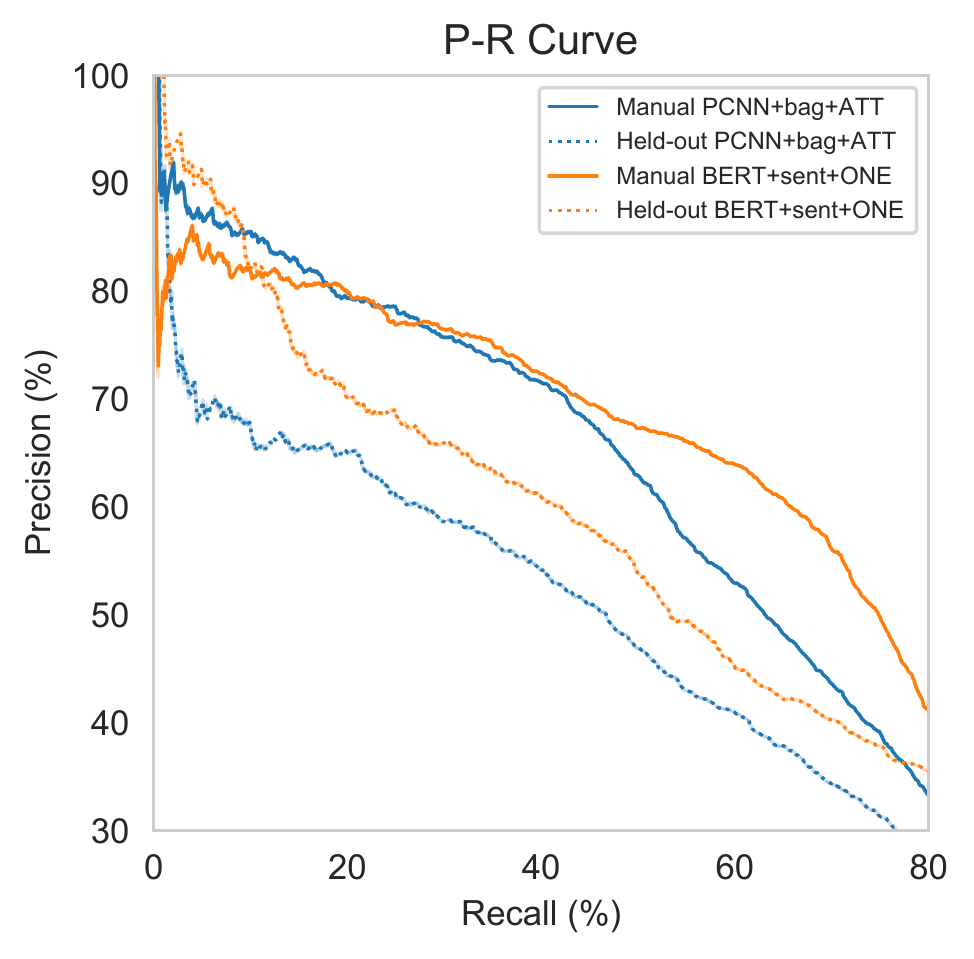}
    \caption{Comparisons of using bag-level manual test (solid lines) and corresponding held-out test (dotted lines). Both absolute and relative scores of models significantly change when taking on human-labeled data.}
    \label{fig:nyt_corr}
    \vspace{-1em}
\end{figure}

Table~\ref{tab:main} shows the main results on \nyt. We also plot P-R curves of selected models in Appendix~\ref{sec:nyt_pr}. 
% and we also plot selected representative models' P-R curves in Figure~\ref{fig:nyt_pr}. Not all models are included in the P-R curves, since many models have close results, and we merge these results for clearer observations.
% We  plot all of them for it is hard to distinguish them then, especially for manual test where lines are closer. 
%From those results, we can make the following observations:
Overall, for all three settings, training pre-trained models in a sentence-level style always perform the best, while applying bag-level training strategies can significantly boost the performance when taking other non-pre-trained encoders. \citet{feng2018reinforcement} observe that bag-level training is not helpful in the sentence-level evaluation, which contradicts our observation. We suspect that it is because \citet{feng2018reinforcement} only manually check a small proportion of test data, leading to a biased result.

More importantly, by comparing the held-out test results to the manual ones, we come to the conclusion that \textbf{manual evaluation matters}: 
auto-labeled and human-labeled test data lead to very different observations. For example, the comparisons between PCNN and RL-DSRE, and BGWA and RESIDE are reversed when taking different evaluations. 
Also, the performance gaps between different models become much smaller when it comes to the manual test. 
Since our manual test set is smaller than the original held-out one (because we did not annotate all N/A sentences), 
%to make a clearer comparison, we take the manual test data and all corresponding with their manual labels and DS labels, 
%%%% This following sentence is particularly hard to address/understand. 
to make a clearer comparison, we evaluate two selected models on the bag-level manual test set and on the corresponding instances in the held-out test set, respectively, 
and we plot the P-R curves in Figure~\ref{fig:nyt_corr}. 
It shows that not only the absolute values of the two measurements differ a lot, but it also affects the relative performances between the models. 
For instance, BERT+sent+ONE shows a considerable advantage over PCNN+bag+ATT at the top predictions on the held-out test set, but it is completely the opposite case at the manual test, where BERT+sent+ONE is even significantly worse than PCNN+bag+ATT. 
It clearly suggests that using the held-out test set cannot well demonstrate the real pros and cons of the models.

\begin{figure}[t]
    \centering
    \includegraphics[width=0.4\textwidth]{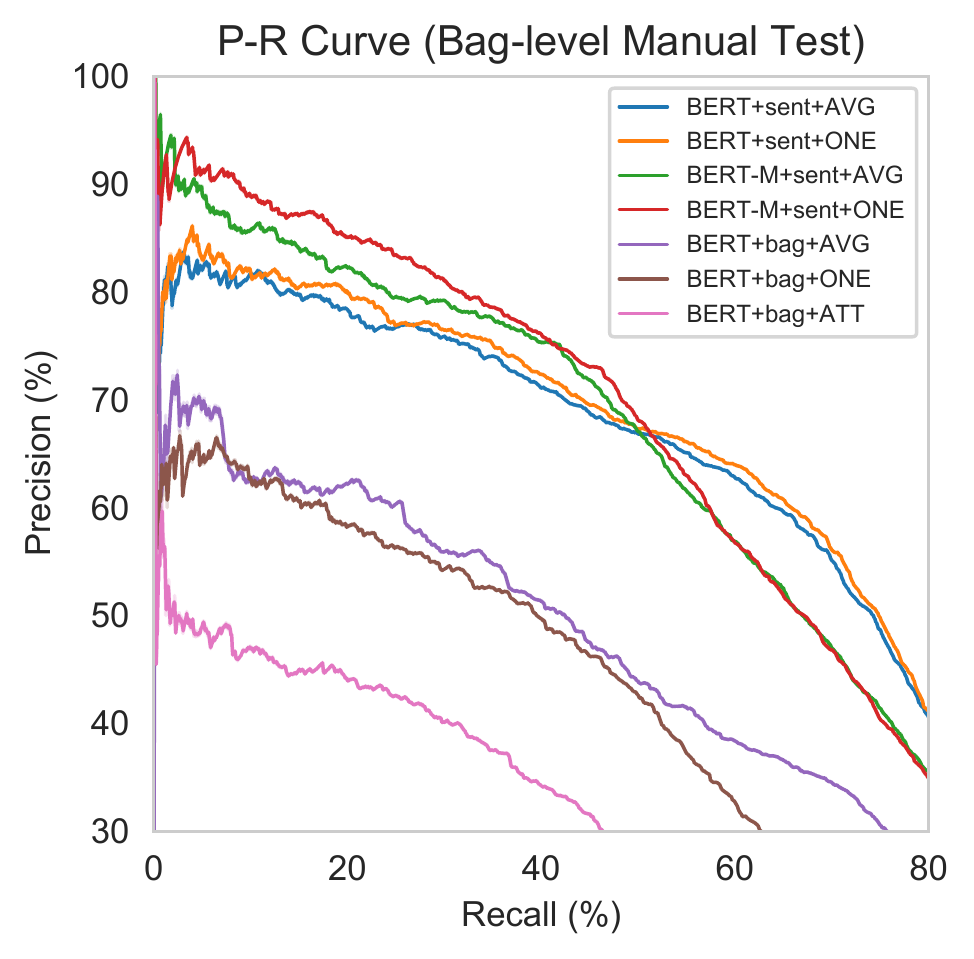}
    \caption{Comparisons of different DS-RE models with BERT on \nyt~bag-level manual test set.}
    \label{fig:nyt_bert}
    \vspace{-0.5em}
\end{figure}

Compared to others, BGWA and RESIDE suffer an extreme change in performance between the held-out and manual evaluations, and we suspect that it is due to the fact that they use entity types as extra information, which leads to overfitting biased heuristics of entities. This further emphasizes the need of using manually-labeled test data in DS-RE.

After checking the manual results, we further identify some interesting observations that have not been clearly demonstrated with the DS evaluation:

\paragraph{Pre-trained Models} 

First of all, BERT-based models have achieved supreme performance across all three metrics. 
To thoroughly examine BERT and its variants in the DS-RE scenario, we further plot their P-R curves with the bag-level manual test in Figure~\ref{fig:nyt_bert}.
It is surprising to see that all bag-level training strategies, especially the ATT strategy which brings significant improvements for PCNN-based models, 
do not help or even degenerate the performance with pre-trained ones. 
%In fact, introducing bag-level training even degenerates the performance a lot. 
This observation is also consistent with that in \citet{amin2020data}, though they only compare BERT+bag+AVG and BERT+bag+ATT. 
We hypothesize the reasons are that 
solely using pre-trained models already makes a strong baseline, since they exploit more parameters and they have gained pre-encoded knowledge from pre-training~\cite{petroni-etal-2019-language}, all of which make them easier to directly capture relational patterns from noisy data; 
and bag-level training, which essentially increases the batch size, may raise the optimization difficulty for these large models.

Another unexpected observation is that, 
though the P-R curve of BERT is far above other models in the held-out test, we identify a significant drop of that in the manual test, as shown in Figure~\ref{fig:nyt_corr} and Appendix~\ref{sec:nyt_pr}. 
By manually checking those errors, we find that most of them are models predicting facts that exist in the KG but are not supported by the text (i.e., false-positive).
For example, \emph{Arthur Schnitzler} was indeed born in \emph{Vienna}, but it is wrong for the model to infer the relation \emph{place of birth} from sentence ``\emph{Authur Schnitzler wrote a story set in Vienna.}''
We assume that it is not only because of the prior knowledge of pre-trained models, but is also due to that BERT can better learn heuristics from entity themselves, as shown in the study of~\citet{peng2020learning} with supervised RE. Considering the data of DS-RE are noisy and in many cases the text does not support the labeled facts, this overfitting-to-heuristic phenomenon can only be more severe.

\begin{figure}[t]
    \centering
    \includegraphics[width=0.48\textwidth]{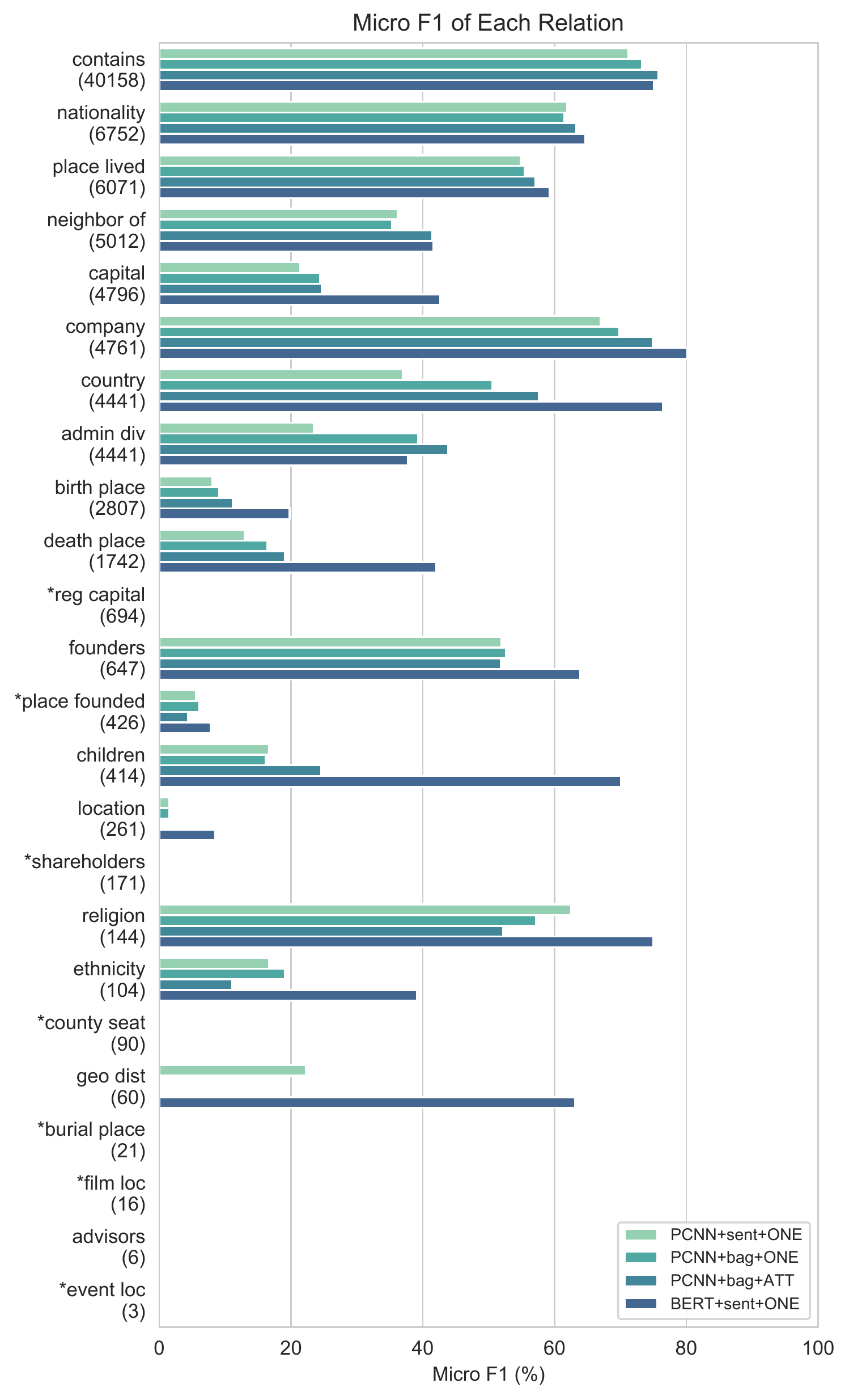}
    \caption{Micro F1 (\%) of different relations on the bag-level manual test of \nyt. $^*$ represents the relation has less than 20 sentences in the test sets. 
    %To save space, we use abbreviations of relation names. 
    Numbers in brackets are numbers of training sentences. %Top-4 relations take more than 80\% of the test sentences.
    }
    \label{fig:rel_f1}
    \vspace{-1.0em}
\end{figure}

%\tianyu{Add BERT+type?}

% \lyk{Remove BERT-M from Table 4, need to add a table or figure comparing BERT and BERT-M}

To verify the assumption and try out a simple solution to alleviate the problem, we take a BERT-M variant (as described in \S\ref{sec:model_baselines}) and show its results in Figure~\ref{fig:nyt_bert}. 
%Though masking entities leads to dramatically drop in the held-out test (which supports that models learn important heuristics from entity names),
%BERT-M variants show significant advantage over BERT at the top predictions on the P-R curve, de 
% it diminishes false-positives in manual test. 
% From Figure~\ref{fig:nyt_bert}, 
We can see that the P-R curves of BERT-M are above those of BERT at the beginning, demonstrating that BERT-M models have higher precisions at those top predictions. Later on, since BERT can extract more facts than BERT-M by fully utilizing information from entity names, BERT-M reduces below BERT. 
From these results, we highlight that \tf{how to handle the false-positives and denoise DS-RE data for pre-trained models} still remains an open and challenging problem. 

% \textbf{how to adopt pre-trained models to handle DS data (especially handle large amounts of noisy instances) still remains an open problem.}

\paragraph{Imbalanced Classes} 

Previous works of DS-RE usually take AUC, micro F1, or P-R curves to measure the abilities of models, which show the overall performance trend averaged on relational facts. 
However, the distribution of training instances across relations is extremely uneven. 
For example, in \nyt~, almost half of the positive instances are \ttt{/location/location/contains}. On the contrary, half of the relations have fewer than $1,000$ sentences. In this case, macro F1 can better show the averaged performance across different relations, without being biased by the majority class. 
Table~\ref{tab:main} demonstrates that even though in most times conclusions of different metrics are consistent, there are cases when models improve micro F1 but degenerate macro F1. 

To further study how models perform on each relation, we plot several representative models and their micro F1 scores for each relation in Figure~\ref{fig:rel_f1}.
We can see that:
(1) The top-4 relations, which account for 80\% test instances, do not vary much in performance with different models, while the difference of performance takes place mostly outside the top-4; 
(2) Some relations even have zero F1 scores, mostly because they have very few training or test sentences.
These results further underscore the importance to look into per-relation scores for DS-RE, and we advocate that later works should \tf{include macro F1} for more comprehensive comparisons.

\subsection{The Results on \wiki}

\input{tables/wiki.tex}

We choose several representative models and further evaluate them on \wiki, as shown in Table~\ref{tab:wiki}. The main observation of results on \wiki~is consistent with that of \nyt---sentence-level pre-trained models perform the best, and using bag-level training helps with non-pre-trained ones---though the overall performance is much higher.   
Another difference is that, in \wiki, AVG performs better than ONE and ATT. 
We think that it is due to the inherent difference in how the two datasets are constructed, especially the difference in the aspect of determining N/A sentences. 
Compared to \nyt, part of the N/A instances in \wiki, instead of indicating no relation between the entities, may correspond to a specific relation that is outside of the dataset ontology. 
% part of the N/A sentences in \wiki~do not have no relations, and may correspond to a specific unknown relations. 
It suggests that when dealing with N/A instances, \tf{considering their latent semantics, rather than simply treating them as one abstract class}, may further benefit RE models.

% Hence, considering the semantic features of N/A instances rather than ignoring them benefits RE models. 

%% file: tables/wiki.tex
%!TEX root = ../acl2021.tex

\begin{table}[t]
    \begin{center}
    \centering
    \small
    %\resizebox{1.0\columnwidth}{!}{%
    \setlength{\tabcolsep}{5pt}
    \begin{tabular}{lcc|ccc}
    \toprule
    Model & Bag & Strategy &  AUC & Micro & Macro\\
    \midrule
    \multirow{5}{*}{PCNN}& - & AVG & 74.1 & 69.1 & 67.1\\
     & - & ONE & 74.0 &	69.1&	66.9\\
     \cmidrule{2-6}
    & \checkmark & AVG & 78.1 & 71.8 & 69.5\\
    & \checkmark & ONE & 76.6	&70.3&	67.7\\
    & \checkmark & ATT & 77.5	&71.2&	68.6\\
    \midrule
    \multirow{5}{*}{BERT}& - & AVG & 90.0 & 83.5 & 82.9 \\
    & - & ONE & 89.8	&83.3&	82.6\\
    \cmidrule{2-6}
   % BERT-M (AVG) & 80.4 & 73.8 & 72.5 \\
   % BERT-M (ONE) & 79.1	& 72.6 & 70.8 \\
    & \checkmark & AVG & 89.9	& 82.7 & 82.0 \\
    & \checkmark & ONE & 88.9	& 81.6 & 81.1 \\
    & \checkmark & ATT & 70.9	& 66.8 & 64.3 \\
    \bottomrule
    \end{tabular}
    %}
    \end{center}
    \caption{Results (\%) on \wiki~of representative models. 
    ``Bag'' indicates bag-level training and ``Micro'' and ``Macro'' represent micro and macro F1 respectively.
    %We report the AUC, Micro F1 (micro) and Macro F1 (macro) scores here.
    }
    \label{tab:wiki}
%\vspace{-1em}
\end{table}

%% file: sections/appendix.tex
%!TEX root = ../acl2021.tex

\section{Relation Ontology Changes of \nyt}
\label{sec:nyt_rel}
%s%%df

\begin{minipage}{\textwidth}
  \vspace{20pt}
  \centering
  \small
  \begin{tabular}{l}
    \toprule
    /location/country/administrative\_divisions\\
    /location/administrative\_division/country\\
    /location/country/capital\\
    (merge) /location/region/capital\\
    \idt /location/fr\_region/capital\\
    \idt /location/cn\_province/capital\\
    \idt /location/in\_state/administrative\_capital\\
    \idt /location/in\_state/legislative\_capital\\
    \idt /location/in\_state/judicial\_capital\\
    \idt /location/it\_region/capital\\
    \idt /location/br\_state/capital\\
    \idt /location/mx\_state/capital\\
    \idt /location/province/capital\\
    \idt /location/us\_state/capital\\
    \idt /location/jp\_prefecture/capital\\
    \idt /location/de\_state/capital\\
    /location/us\_county/county\_seat\\
    /location/neighborhood/neighborhood\_of\\
    /location/location/contains\\
    (merge) /business/location \\
    \idt /business/company/locations\\
    \idt /sports/sports\_team/location\\
    \idt /broadcast/producer/location\\
    /business/company/founders\\
    /business/company/place\_founded\\
    /business/company/major\_shareholders\\
    /business/company/advisors\\
    /business/person/company\\
    /people/person/place\_of\_birth\\
    /people/person/religion\\
    /people/person/nationality\\
    /people/person/place\_lived\\
    /people/person/ethnicity\\
    /people/person/children\\
    /people/deceased\_person/place\_of\_death\\
    /people/deceased\_person/place\_of\_burial\\
    /people/ethnicity/geographic\_distribution\\
    /time/event/locations\\
    /film/film/featured\_film\_locations\\
    N/A\\
    (delete) location/country/languages\_spoken\\
    (delete) base/locations/countries/states\_provinces\_within\\
    (delete) business/shopping\_center\_owner/shopping\_centers\_owned\\
    (delete) business/shopping\_center/owner\\
    (delete) business/business\_location/parent\_company \\
    (delete) business/company\_advisor/companies\_advised \\
    (delete) people/profession/people\_with\_this\_profession \\
    (delete) people/person/profession\\
    (delete) people/place\_of\_interment/interred\_here \\
    (delete) people/ethnicity/included\_in\_group \\
    (delete) people/family/members \\
    (delete) people/family/country \\ 
    (delete) broadcast/content/location \\
    (delete) film/film\_festival/location \\
    (delete) film/film\_location/featured\_in\_films \\
    \bottomrule
    \end{tabular}
    %}
    %\end{center}
    \captionof{table}{\nyt~relation ontology. All those relations are from FreeBase. ``(merge)'' means a merge relation in our version of \nyt~and it is followed by relations merged from the original dataset. ``(delete)'' indicates that this relation is discarded in our version because there are no instances in the training or the test sets.}
\end{minipage}

%\bibliographystyle{acl_natbib}
%\bibliography{acl2021}

\clearpage

\section{P-R Curves for \nyt}
\label{sec:nyt_pr}

  \begin{minipage}{\textwidth}
    \vspace{20pt}
    \centering
    \includegraphics[width=0.32\textwidth]{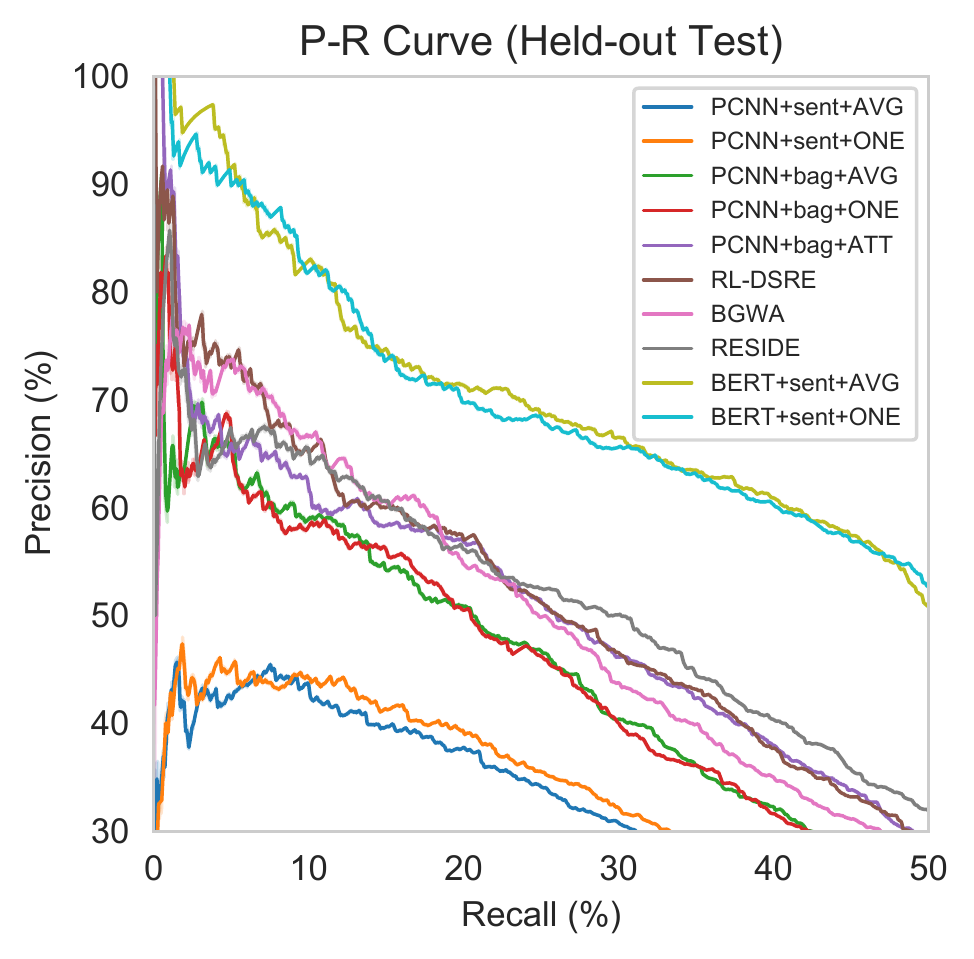}
    \includegraphics[width=0.32\textwidth]{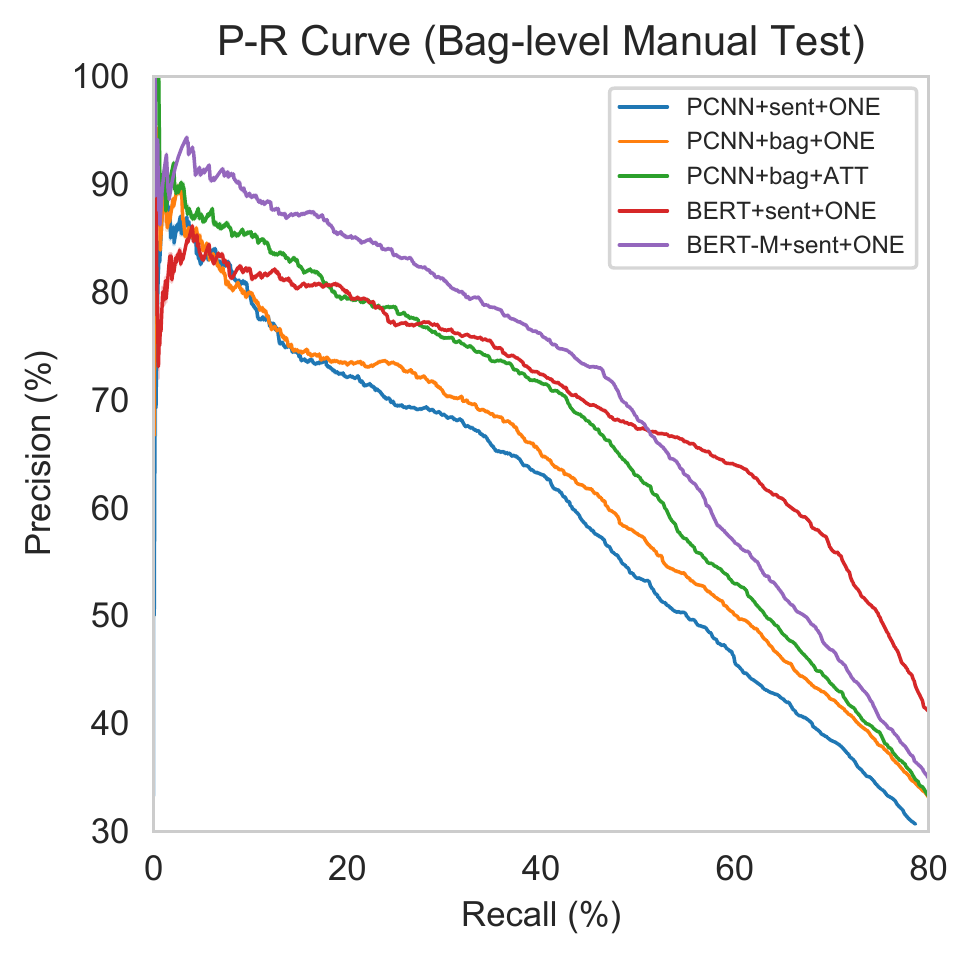}
    \includegraphics[width=0.32\textwidth]{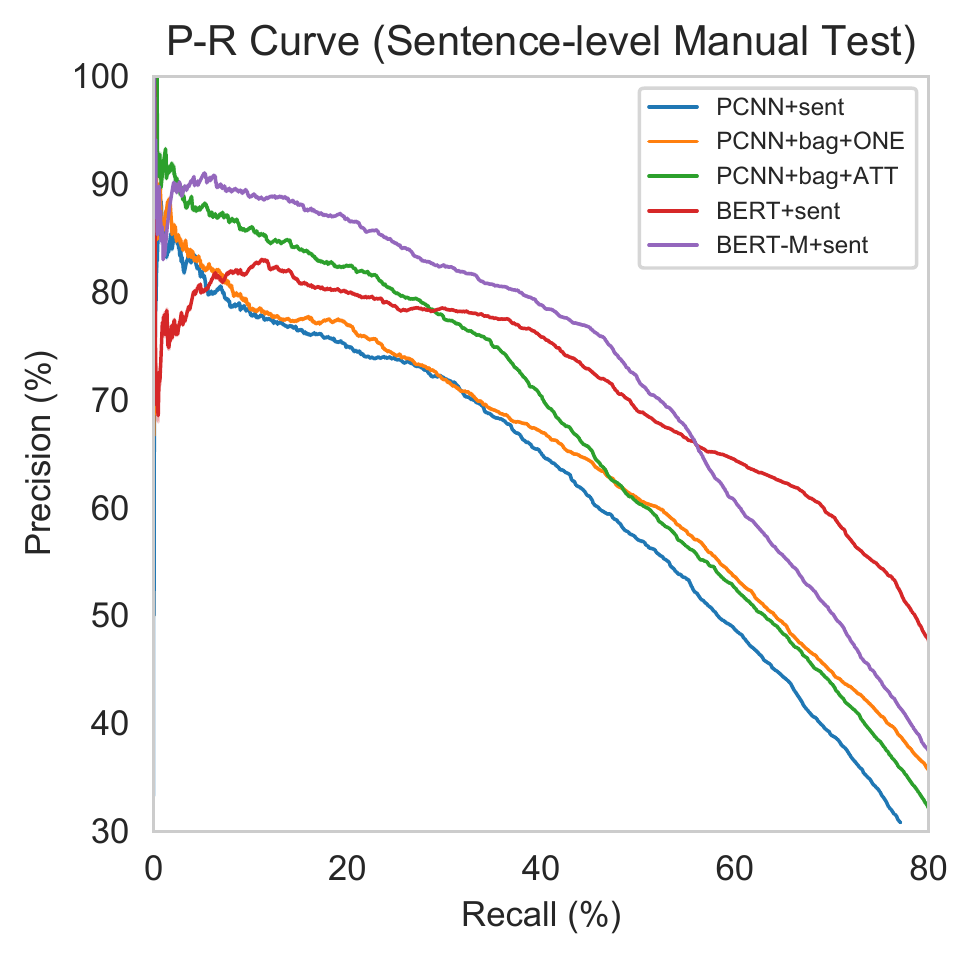}
    \captionof{figure}{P-R curves of representative models in held-out test, bag-level manual test and sentence-level manual test of \nyt. Note that the scales of X-axis are not the same in the three figures.}
    \label{fig:nyt_pr}
  \end{minipage}